\pdfoutput=1
\documentclass[11pt]{article}

\usepackage{acl}
\usepackage{kotex}

\usepackage{times}
\usepackage{latexsym}
\usepackage{booktabs}       % professional-quality tables
\usepackage{amsfonts}       % blackboard math symbols
\usepackage{nicefrac}
\usepackage{amsmath}
\usepackage{multirow}

\usepackage{graphicx}
\usepackage{subfigure}
\usepackage{tabularx} % 표 너비 줄이기
\usepackage{relsize} % 폰트 크기 설정

\usepackage[T1]{fontenc}
\usepackage[utf8]{inputenc}

\usepackage{microtype}

\title{Generation-Based and Emotion-Reflected Memory Update: Creating the KEEM Dataset for Better Long-Term Conversation}

\author{Jeonghyun Kang$\thanks{\,\, Equal contribution}$ \\
    \\ \And
  Hongjin Kim$^{\ast}$ \\
  Konkuk University \\
  \texttt{\{jeonghyun97, jin3430, nlpdrkim\}@konkuk.ac.kr} \\ \And Harksoo Kim$\thanks{\,\, Corresponding author}$}

\begin{document}
\maketitle
% \begin{abstract}
% In this work, we introduce the Keep Emotional and Essential Memory (KEEM) dataset, a novel, generation-based approach that integrates both emotion and causality into memory updates for long-term conversation. Unlike previous datasets that rely on simple accumulation or operation-based methods for memory updates, our KEEM dataset employs a dynamic generation of integrative memories. This approach not only retains essential information but also incorporates emotion and causality, facilitating deeper empathy and understanding in open-domain conversational systems.\footnote{Our code and KEEM dataset will be available in Github.}
% \end{abstract}
\begin{abstract}
In this work, we introduce the Keep Emotional and Essential Memory (KEEM) dataset, a novel generation-based dataset designed to enhance memory updates in long-term conversational systems. Unlike existing approaches that rely on simple accumulation or operation-based methods, which often result in information conflicts and difficulties in accurately tracking a user’s current state, KEEM dynamically generates integrative memories. This process not only preserves essential factual information but also incorporates emotional context and causal relationships, enabling a more nuanced understanding of user interactions. By seamlessly updating a system’s memory with both emotional and essential data, our approach promotes deeper empathy and enhances the system’s ability to respond meaningfully in open-domain conversations.
% \footnote{Our code and KEEM dataset will be available in Github.}
\end{abstract}

\section{Introduction}
Open-domain conversational systems have increasingly achieved human-like interaction capabilities. Recent advancements have led to the development of long-term conversation chatbots and datasets \cite{xu2022beyond}. Despite these significant strides, the study of memory management in long-term conversations has not been adequately addressed. In real-life conversations, a user's state or conditions frequently change, necessitating dynamic memory adjustments. Previous methods have primarily relied on accumulating new memories while retaining previous ones, a strategy that inadequately addresses the need for revising or deleting outdated information. Furthermore, in multi-session conversations, as dialogues extend, the requirement for expanded memory capacity intensifies. Although some studies \cite{shuster2022blenderbot, ma2021one, xu2022long} have suggested a simplistic approach of deleting old memories to make room for new ones, this method fails to recognize the necessity of updating previous memories based on new information, highlighting the need for a more sophisticated memory management solution. \\
% 기존의 메모리를 새로 진행되는 대화에 나타난 정보와 비교해 최신 정보로 유지하는 메모리 업데이트는 업데이트된 메모리를 기반으로 상대에게 근황을 묻고 상대가 했던 말을 기억해줌으로써 long-term conversation에서 사용자가 더 높은 친밀감을 느끼고 이로 인해 대화에 더 적극적으로 참여할 수 있도록 한다. 또한 최신 정보로 유지하기 위해 기존의 메모리를 삭제 또는 수정하기 때문에 메모리 capacity문제를 보완할 수 있고 사용자의 최신 상태를 추적하는 데에도 용이하다. (어떤 시스템에서 더욱 필요할지 생각해보기) 
% Updating memory according to information from new dialogue by comparing the previous memory can encourage users to engage in converse because it can ask about the recent condition and remember what users said based on updated memory. In addition, it can alleviate memory capacity and help track the user's recent condition since it \textit{DELETE}s or revises the previous memory to retain a new state.
 % GPT revised version
\indent Updating memory based on information from new dialogues, through comparison with previous memories, can significantly enhance user engagement. This approach allows conversational systems to inquire about recent conditions and reference past conversations accurately, based on updated memories. Moreover, this method contributes to efficient memory capacity management by deleting or revising outdated memories to reflect the latest user state. Such dynamic memory management not only supports maintaining a concise and relevant memory database but also facilitates accurate tracking of the user's current conditions, thereby improving the conversational system's responsiveness and personalization.
\begin{figure*}[t]
    \centering
    \includegraphics[width=1.0\textwidth]{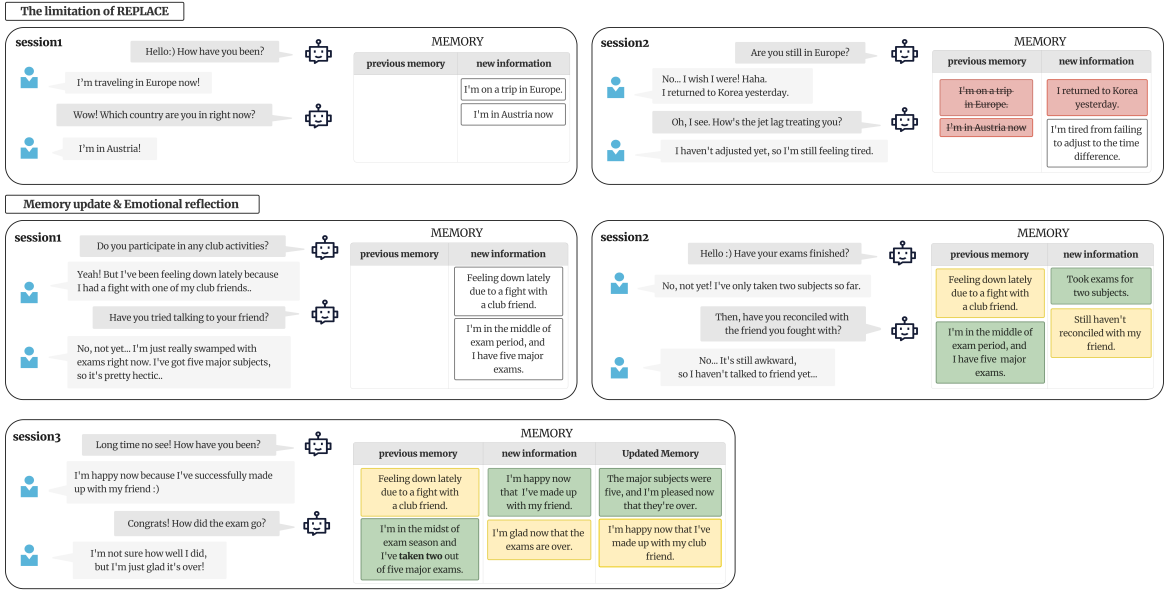}
    \caption{Limitations of previous method and example of memory update.}
    \label{figure_limitations_replace}
\end{figure*}
% GPT revised version
Despite the importance of memory management, it has been relatively understudied, possibly due to the scarcity of relevant datasets. \citet{bae2022keep} introduced a method and dataset for memory updating named \textit{CareCallmem}, which classifies the relationships between sentence pairs into four operations: \textsc{PASS}, \textsc{APPEND}, \textsc{DELETE}, and \textsc{REPLACE}. These operations enable the identification and removal of information that could cause confusion or redundancy in subsequent conversations. The \textsc{DELETE} operation eliminates both the previous and new memory when the information is no longer needed, while \textsc{REPLACE} retains only the new memory in cases of conflict between the old and new information. However, these operations present significant challenges in memory management. For instance, determining the appropriate operation can be ambiguous, as illustrated when comparing sentences like ``I have a cold and am taking medicine'' with ``I am completely cured of my cold,'' leading to confusion about the correct operation to apply. Furthermore, using \textsc{DELETE} in situations where both sentences, such as ``I have COVID'' and ``I am completely cured of COVID,'' are eliminated, the system loses the ability to remember that the user had COVID. Similarly, with \textsc{REPLACE}, if the system updates from ``I am traveling in Europe'' to ``I am back in Korea,'' it fails to retain the information that the user was previously in Europe (See Figure \ref{figure_limitations_replace}). In daily conversations, user experiences and episodes unfold over time. For instance, processing two memories, ``I have 4 subjects left for the exam'' and ``I took the exam for 1 subject,'' using \textsc{DELETE} or \textsc{REPLACE} operations proves inadequate. To address these challenges, we introduce a generation-based memory update dataset designed to preserve essential information for long-term conversation systems. Our approach aims to generate a new sentence that retains critical information when confronted with sequential memories, such as transitioning from ``I am traveling in Europe'' to ``I am back in Korea.'' Instead of merely selecting or removing, our system generates an integrative memory, such as ``I traveled to Europe,'' ensuring the preservation of essential user information. \\
% GPT made it shorter
% Despite the critical role of memory management in conversational AI, it remains underexplored, likely due to a lack of focused datasets. \citet{bae2022keep} proposed \textit{CareCallmem}, a dataset that classifies sentence relationships into four operations (\textit{PASS}, \textit{APPEND}, \textit{DELETE}, \textit{REPLACE}) for memory updating. While these operations aim to manage conversational memory by removing or updating irrelevant or conflicting information, they face challenges in preserving essential user details. For example, \textit{DELETE} might remove all mentions of an event like recovering from an illness, losing valuable context, whereas \textit{REPLACE} could omit past activities, such as travel when updating to current states, as shown in Fig. \ref{figure_limitations_replace}. To overcome these limitations, we introduce a generation-based dataset that enhances memory management by dynamically generating integrated updates that retain critical information across conversations. This approach ensures that essential details, like the progression from having pending exams to completing them, are seamlessly incorporated into the memory, thereby maintaining continuity and context in long-term interactions. \\

\indent Existing multi-session chat (MSC) datasets often focus on summarizing events or episodes, neglecting the emotions related to these events or failing to account for the causes of the user's emotions. Emotion plays a pivotal role in human-like conversational systems. Systems that reference emotions without their underlying causes offer only superficial empathy. In contrast, acknowledging both emotions and their causes facilitates a deeper understanding and empathy \cite{gao2021improving}. This becomes particularly valuable when users express negative emotions, enabling the system to offer solutions or advice, with potential applications in mental health care \cite{wang2023generative}. To address this gap, our memory update dataset incorporates both emotion and its causes from dialogues, as shown in Figure \ref{figure_limitations_replace}. Finally, we introduce the Keep Emotion and Essential Memory (KEEM) dataset, which is generation-based and uniquely integrates emotional context and causality. To sum up, our main contributions are as follows:

\begin{enumerate}
    \item We propose a novel generation-based memory update dataset that enhances long-term conversational systems by dynamically generating integrative memories to retain essential information, addressing the limitations of existing \textsc{DELETE} and \textsc{REPLACE} operations.
    \item Our work introduces the Keep Emotion and Essential Memory (KEEM) dataset, the first of its kind to integrate both emotion and causality into memory updates facilitating deeper empathy and understanding in conversational AI.
    \item We demonstrate the quality and applicability of our dataset in improving user engagement and system responsiveness in multi-session conversations.
\end{enumerate}
%, particularly highlighting its potential in mental health care by providing solutions or advice based on a nuanced understanding of users' emotions and their causes.

 % \begin{figure}[t]
%     \centering
%     \includegraphics[width=0.5\textwidth]{figure/subfigure2.jpg}
%     \label{figure_update_example}
%     \caption{Example of memory update.}
% \end{figure}
\section{Related Work}
To support the study of long-term conversations, \citet{xu2022beyond} developed an MSC dataset, which comprises multiple dialogue sessions. Each session in the MSC dataset is accompanied by a summary that serves as a memory for subsequent dialogue sessions. However, while the MSC dataset facilitates the continuation of dialogues across sessions, it does not explicitly address the process of memory updates between these sessions. \\
\indent \textit{CareCallmem} \cite{bae2022keep} represents the first Korean memory update dataset, developed from the \textit{CareCall} dataset, which was designed for a role-specific open-domain dialogue system targeting elderly individuals living alone. The \textit{CareCall} system's dialogues predominantly center on the users' well-being, emphasizing health and meals, which narrows the diversity of dialogues within the dataset. Moreover, the summaries within the \textit{CareCall} dataset often omit significant details, tending to highlight events at the expense of the users' emotions or thoughts about these events. This approach to summarization leads to brief summaries that overlook the nuanced points of events and fail to capture the full context, potentially resulting in inaccurate memory updates in subsequent dialogues (See example in Appendix \ref{Carecallmem_examples}).

Recent advancements in empathetic dialogue (ED) research have expanded on the ED dataset initially introduced by \citet{rashkin-etal-2019-towards}. \citet{sabour2022cem} have developed a model that boosts cognitive empathy by using common sense to better understand and empathize with users' situations. Furthermore, \citet{zhou2022case} introduces a technique that merges cognitive insights with emotional reactions through their CASE model, highlighting the synergy between cognitive and emotional elements. Nonetheless, these studies primarily concentrate on crafting empathetic responses using the dataset that is based on English. Recently, the Korean cultural dialogue dataset was constructed \cite{jin-etal-2024-kobbq}; however, this dataset does not focus on memory updates for long-term dialogue.
\section{KEEM Dataset Creation}
\begin{figure*}[t]
\centering
\includegraphics[width=1.0\textwidth]{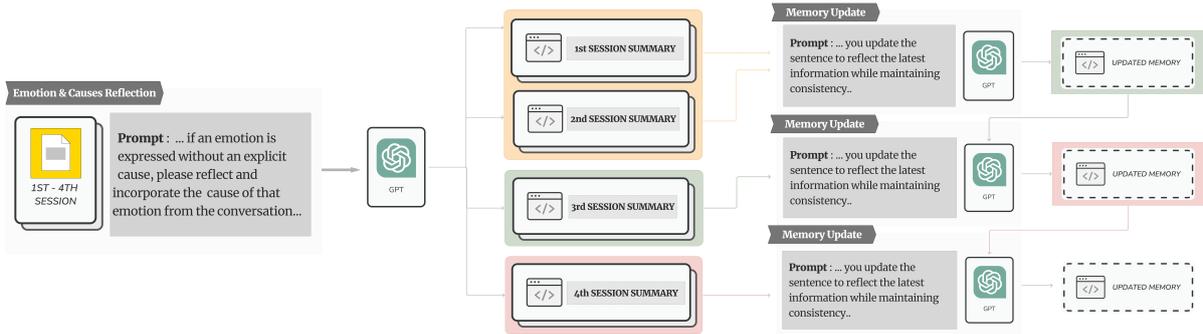}
\caption{Process of KEEM dataset creation.}
\label{fig:framework}
\end{figure*}
To diversify dialogue topics, we opt for the Korean MSC (KMSC) dataset from AI Hub\footnote{https://www.aihub.or.kr/} instead of the CareCall dataset. The KMSC dataset encompasses 13 themes, such as `individuals and relationships', `education', `climate', and `beauty and health', among others (for more details on the KMSC dataset, see Appendix  \ref{sec_topic}). It comprises four sessions, each with a summary. However, it is not designed with memory updates in mind; summaries from each session are merely appended and used as the basis for subsequent dialogues' memories. Furthermore, similar to other datasets, the KMSC summaries often lack information on the user's emotions or their causes, with some containing emotions but omitting their origins(utterance: ``I work as a designer. Recently, I sometimes feel embarrassed because of past work mistakes.'' -> summary: ``I’m a designer and I sometimes feel embarrassed these days.''). \\
\indent To construct our KEEM dataset, which integrates both emotion and causality into generation-based memory updates, we create new summaries using dialogues from the KMSC dataset and ChatGPT 4.0 \cite{achiam2023gpt}. ChatGPT has demonstrated significant capabilities in various NLP tasks, including dataset creation \cite{wang2023selfinstruct}, with multiple studies validating its effectiveness and efficiency \cite{cegin-etal-2023-chatgpt}. We leverage ChatGPT to first incorporate users' emotions and their causes into the summaries. Subsequently, ChatGPT is instructed to generate an updated summary that combines the previous and current sessions' summaries. Figure \ref{fig:framework} shows the framework of our KEEM dataset creation. Table \ref{tab:data_statistics} shows the statistics of the KMSC and our KEEM datasets. In Table \ref{tab:data_statistics}, note that 'Session 1-$n$' refers to data comprising $n$ sessions. For example, 'Session 1-4' indicates data encompassing four sessions, not the fourth session.
% The statistics of our KEEM dataset are in Appendix ~\ref{statistics_datasets}.
\subsection{Emotions \& Causes Reflection \label{emotion_causes}}
% Given the session's dialogue $D = \{u_1, u_2, ..., u_n\}$, where $n$ denotes the number of utterances in a session, and the session's summary $S = \{s_1, s_2, ..., s_m\}$, where $m$ denotes the number of the summary sentences, we instruct ChatGPT to reflect the user's emotions and their cause when any cause appears from $D$ into $S$. We try three types of prompting methods for reflecting emotions and their causes. 1) First, we try the instructions in both English and Korean since our dialogues are Korean. We input English and Korean instructions on 100 samples, then we manually evaluate the results from each instruction (For result examples, please see Appendix \ref{emotionCausesExample}. As a result, the Korean instruction is better so we select it for all datasets. 2) Second, we try the reasoning prompt which gives the candidates of emotion in the instruct. For the candidates of the emotions, we analyze and observe the emotions that appeared in the KMSC dataset, such as pleasure, sadness, happiness, etc (For all the emotions, please see Appendix \ref{emotionCausesExample}). We sample 100 examples and compare results with and without the candidates of emotions. As a result, when giving the candidates of the emotions ChatGPT generates a better summary while reflecting the emotions and their causes. 3) Third, we try to give the few-shot examples for eliciting ChatGPT's capability, and it shows better results. Finally, we select the Korean instruction and give the candidates the emotions and few-shot examples for our final prompt.
% GPT revised version
Given a session's dialogue $D$ and the session's summary $S$, we instruct ChatGPT to incorporate the user's emotions and their causes into $S$, drawing from any cues present in $D$. We explore three prompting methods to effectively reflect emotions and their causes: 1) Initially, we test instructions in both English and Korean, given the dialogues are in Korean. We apply these instructions to 100 samples and conduct a manual evaluation of the outcomes from each language setting (For examples of the results, see Appendix \ref{appendix:emotions & causes reflection}). The Korean instructions yield superior results in reflecting dialogue content and updating whole dialogue contexts; thus, we have adopted them for the entire dataset. 2) We then experiment with a reasoning prompt that include potential emotions observed in the dialogues, such as pleasure, sadness, and happiness, among others. We assess 100 examples comparing the effectiveness of prompts with and without specified emotion candidates. The prompt without emotion candidates generate summaries that more accurately reflected emotions and their causes. 3) Lastly, we employ a few-shot approach, providing ChatGPT with specific examples to enhance its performance. This method demonstrates improved outcomes in capturing the nuances of emotions and their causes. Consequently, we select the Korean instructions without emotion candidates, incorporating few-shot examples, as our final prompting strategy.

\begin{table*}[t]
\centering
\relscale{0.84} % 폰트 크기 배율
\setlength\tabcolsep{5pt} % 표의 컬럼 간격 줄이기, 기본값: 6pt
\renewcommand{\arraystretch}{1.1}
\begin{tabular}{l|rrr|rrr}
\toprule
\multirow{2}{*}{\textbf{Attribute}} & \multicolumn{3}{c|}{\textbf{KMSC}} & \multicolumn{3}{c}{\textbf{KEEM}} \\
 & \textbf{Session 1-2} & \textbf{Session 1-3} & \textbf{Session 1-4} & \textbf{Session 1-2} & \textbf{Session 1-3} & \textbf{Session 1-4} \\
\midrule
Total Episodes                        & 40,000 & 20,000 & 20,000 & 2,006 & 1,560 & 1,005   \\
Total Utterances                      & 980,919 & 731,705 & 953,405 & 61,354 & 70,974 & 59,847 \\
Total Memory sentences                & 567,176 & 395,808 & 498,497 & 33,972 & 31,846 & 23,148 \\
Avg. Length of Utterances             & 16.98 & 17.13 & 17.22 & 17.02 & 17.26 & 17.32 \\
Avg. \# of Utterances           & 30.65 & 45.73 & 59.59 & 30.58 & 45.49 & 59.54 \\
Avg. \# of Memory Sentences     & 17.72 & 24.73 & 31.15 & 16.93 & 20.41 & 23.03 \\
\bottomrule
\end{tabular}
\caption{\label{tab:data_statistics}
Statistics of the original KMSC dataset and KEEM dataset. Avg. denotes the average.
}
\end{table*}

\subsection{Memory Update} \label{Sec_memory_update}
In the context of memory updates, given the existing memory (i.e., previous sessions' summaries) $M$ and the current session's summary $S$ (or the current session's dialogue $D$), our objective is to generate an updated memory $M^U$. Consistent with Section \ref{emotion_causes}, we employ Korean instructions and a few-shot approach. We explore two main prompting strategies for effective memory updating: 1) We initiate by comparing the outcomes of providing ChatGPT with either the current session's summary $S$ or dialogue $D$, applying these instructions to 100 samples for a manual evaluation. Given the similarity in results and to minimize costs, we opt for using the summary $S$, as its brevity translates to lower ChatGPT API usage costs. 2) We experiment with a reasoning prompt aimed at facilitating ChatGPT's handling of multiple sentences in both $M$ and $S$, which can be challenging due to the need to correlate summary sentences with the corresponding memory. Initially, we direct ChatGPT to match summary sentences with related memory sentences and then generate the updated memory $M^U$. This approach is tested both as a single integrated instruction and through separate instructions. However, neither method outperformed direct instructions for memory updating (for examples, see Appendix \ref{Appendix_memory_update_examples}). Therefore, we proceed by simply providing ChatGPT with $M$ and $S$, instructing it to update the memory. To ensure the relevance and accuracy of updated memory, we conduct a final verification step using ChatGPT. We input all session dialogues up to the current time step alongside updated memory, instructing the model to verify whether the updated memory accurately reflects the content of the dialogues. If updated memory does not pass this final verification, it is excluded from the KEEM dataset. Post-verification, the refined memory is designated as $M^U$.

\section{Experiments}
\subsection{Implementation Details}
% We used the ChatGPT 4.0 (gpt-4-1106-preview) API to create our KEEM dataset. We set the temperature to 0.0. Both top-p and top-n were configured to 1. We evaluate our KEEM datasets with long-term chatbot models using one NVIDIA A100 80GB GPU for all experiments.
% GPT revised
We utilize the ChatGPT 4.0 (gpt-4-1106-preview) API to create the KEEM dataset, setting the temperature parameter to 0.0. Both top-p and top-n settings are configured to 1 to ensure deterministic output. The evaluation of our KEEM datasets is conducted on one NVIDIA A100 80GB GPU, which supports all experiments involving long-term chatbot models.
\subsection{Human Evaluator Details \label{human_evaluator}}
% For human evaluators, we employ five Korean native speakers who are familiar with chatbot systems and do not know about how our KEEM and \textit{CareCallmem} datasets were created.
% GPT revised version
For the manual evaluation process, we engage five native speakers familiar with chatbot systems as human evaluators. These individuals are not informed about the development processes of either the KEEM or the \textit{CareCallmem} datasets.
\begin{table}[t]
\centering
\small
\begin{tabular}{lcc}
\toprule
Memory         & E.\&C. Ref. rate & Score \\
\midrule
KMSC memory     & 35\%                  & 1.18  \\
KEEM memory &   93\%                                     & 1.90 \\
\bottomrule
\end{tabular}
\caption{Results of manual evaluation of emotion and cause reflection. \textit{E.\&C. Ref. rate} denotes the percentage in which emotions and their causes appear in memory.}
\label{Table_emotions_causes}
\end{table}
% To validate the effectiveness and quality of our KEEM dataset, we conducted various evaluations and experiments. First of all, we evaluated our updated summary in the KEEM dataset, on emotion's causes reflection and memory update. We manually evaluate whether the emotions' causes are reflected in the memory. We set the scale of the score from 0 to 2. A score of 0 implied that the memory does not reflect any emotions and their causes. A score of 1 suggested that the emotions are reflected but their causes are not. A score of 2 indicated both the emotions and their causes are reflected. We sampled 50 sessions and their memory to assess the reflection of emotion and its cause. Table \ref{Table_emotions_causes} shows the results that manually evaluated scores on emotion and its cause reflection. As shown in Table ~\ref{Table_emotions_causes}, our updated memory much better keeps the user's emotion and its cause compared to raw memory. This result implies that using our updated memory can help better empathy with users in subsequent conversations.
% GPT revised version
\subsection{Evaluation of Emotion and Cause}
To validate the effectiveness and quality of our KEEM dataset, we conduct comprehensive evaluations and experiments focusing on the reflection of emotion's causes in the updated summaries. We manually assess whether the causes behind emotions are accurately integrated into the memory updates (Details for the human evaluator are in Section \ref{human_evaluator}). The evaluation is based on a scale ranging from 0 to 2, where a score of 0 indicates no reflection of emotions or their causes, a score of 1 signifies that only emotions are reflected without their causes, and a score of 2 denotes that both emotions and their causes are effectively captured. We analyze 50 sessions and their corresponding memories to evaluate the accuracy of emotion and cause reflection. The results, as detailed in Table \ref{Table_emotions_causes}, demonstrate that our updated memory significantly improves the retention of users' emotions and their causes compared to the raw memory. This enhancement suggests that using our updated memory in conversational AI systems could facilitate deeper empathy in subsequent interactions with users. \\
\begin{table}[t]
    \centering
    \small
    \begin{tabular}{ccc}
    \toprule
        Session & Update Need Rate & Score  \\
        \midrule
        Session 1-2 & 12\% & 1.75  \\
        \midrule
        Session 1-3 & 31\% & 1.64  \\
        \midrule
        Session 1-4 & 21\% & 1.62  \\
        \bottomrule
    \end{tabular}
    \caption{Results of manual evaluation of updated memory. Note that 'Session 1-$n$' refers to data comprising $n$ sessions, rather than indicating the session number. }
    \label{Table_updated_memory}
\end{table}
% Next, to verify our updated memory, we manually evaluated whether the update was accurately performed. The evaluation was based on a scale ranging from 0 to 2, where a score of 0 indicated no memory update occurred, a score of 1 signified that some update was performed, and a score of 2 denoted that the update was accurately performed. We analyzed 100 samples for each session. As shown in Table \ref{Table_updated_memory}, the need for memory updates increases as with session goes on. Moreover, the result shows that the updated memory performed highly accurate memory updates and suggests that using our updated memory in long-term conversational systems could help to track the user's state or conditions.
% GPT revised version
\subsection{Evaluation of Memory Update}
\indent To further validate the accuracy of our memory updates, we conduct a manual evaluation based on a scale from 0 to 2. A score of 0 indicates that no update is performed on the memory, a score of 1 suggests that a partial update is made, and a score of 2 denotes a fully accurate update. This assessment is carried out on 100 samples for each session. Even in long-term conversations, memory updates are not always required after each session. Therefore, we measured the rate at which memory updates are needed across the entire conversation dataset. As depicted in Table \ref{Table_updated_memory}, the necessity for memory updates tends to escalate as the number of sessions increases. The results reveal that our updated memory achieves highly accurate updates, indicating that employing our updated memory in long-term conversational systems could effectively maintain and track the user’s evolving state or conditions. \\
% Finally, to compare our KEEM dataset with the previous update methods, \textit{CareCallmem}, we updated 50 memories in the KMSC dataset by using the methods of \textit{CareCallmem} dataset. To fair comparison, we instruct \textit{DELETE, REPLACE, APPEND}, and \textit{PASS} operations to update the memory in the KMSC dataset using ChatGPT. Then, we conducted a manual evaluation based on a scale from 0 to 2 same manner as the manual evaluation of our updated memory. In addition, we request the evaluators and ChatGPT which memory is better between our updated memory and memory updated through \textit{CareCallmem} manner.
% GPT revised version
\subsection{Comparative Evaluations of Memory Update Methodologies}
\begin{table}[t]
\centering
\small
\begin{tabular}{lccc}
\toprule
\multirow{2}{*}{Updated Memory}                        & \multirow{2}{*}{Score} & \multicolumn{2}{c}{Voting} \\
\cline{3-4} \\[-0.95em]
                                      &       & Human       & ChatGPT      \\
                                      \midrule
KEEM                                  & 1.86     & 76\%           & 81\%            \\
\midrule
\textit{CareCallmem} & 1.60     & 24\%           & 19\%           \\
\bottomrule
\end{tabular}
\caption{Comparative results of memory update methods between the KEEM dataset and the \textit{CareCallmem} dataset.}
\label{Comparison_KEEM_CareCallmem}
\end{table}
\subsubsection{Automatic \& Human Evaluation} \label{sec_automatic_human_eval}
\begin{figure*}[t!]
\centering
\includegraphics[width=\textwidth]{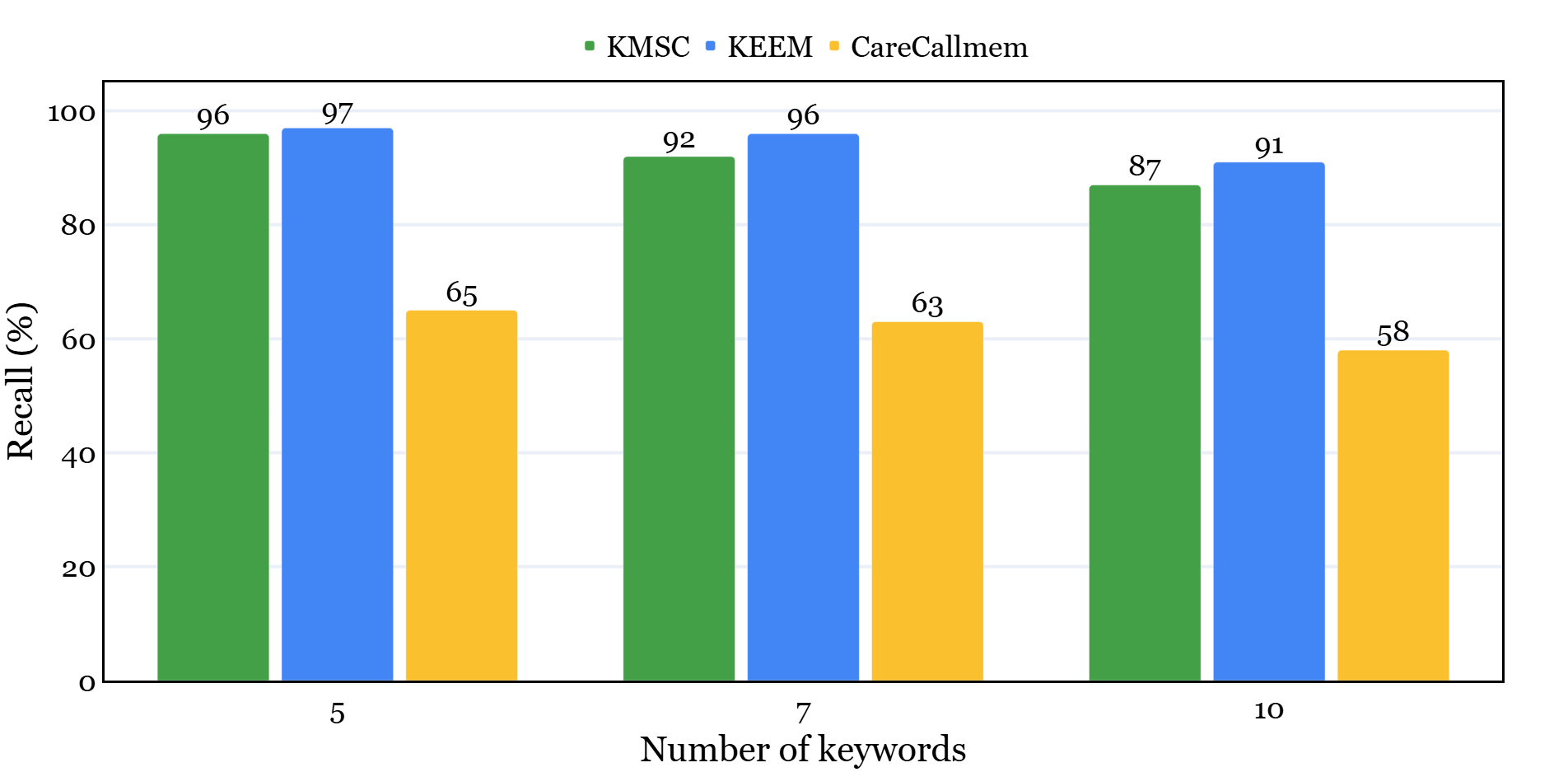}
\caption{Results of keyword coverage across the different memory update methodologies.}
\label{tab_keyword_coverage}
\end{figure*}
\indent To benchmark our KEEM dataset against the existing update method, especially the \textit{CareCallmem} dataset, we apply \textit{CareCallmem}'s methods to update 50 memories in the KMSC dataset. For a fair comparison, we utilize ChatGPT 4.0 API to execute the \textsc{DELETE}, \textsc{REPLACE}, \textsc{APPEND}, and \textsc{PASS} operations for memory updates within the KMSC dataset. Detailed instruction is provided in Appendix \ref{appendix_carecall_instruction}. Subsequently, a manual evaluation is conducted using a scale from 0 to 2, mirroring the assessment criteria used for our memory updates. Moreover, we ask evaluators and ChatGPT to determine which method produces better memory updates: our approach or the \textit{CareCallmem} method.
% Table \ref{Comparison_KEEM_CareCallmem} shows the human and ChatGPT evaluations' results. As shown in Table \ref{Comparison_KEEM_CareCallmem}, our KEEM dataset and \textit{CareCallmem} showed strong performance on human evaluation, and our KEEM dataset outperformed the \textit{CareCallmem} dataset. Moreover, in the voting evaluation of both human and ChatGPT, our KEEM dataset significantly outperformed the \textit{CareCallmem} dataset. It demonstrates that our method of memory update can give a much more comprehensive and informativeness updated memory to the user.
% GPT revised
Table \ref{Comparison_KEEM_CareCallmem} presents the results of evaluations conducted by both human evaluators and ChatGPT. As indicated in Table \ref{Comparison_KEEM_CareCallmem}, both our KEEM dataset and the \textit{CareCallmem} dataset demonstrate strong performance in human evaluations, with our KEEM dataset achieving superior results compared to the \textit{CareCallmem} dataset. Furthermore, in the combined voting evaluation conducted by humans and ChatGPT, the KEEM dataset significantly outperforms the \textit{CareCallmem} dataset. These results underscore the efficacy of our memory update methodology, showcasing its ability to provide more comprehensive and informative updates to users.

\subsubsection{Informativeness \& Conflicts Evaluation}
\begin{figure*}[t]
\centering
\includegraphics[width=\textwidth]{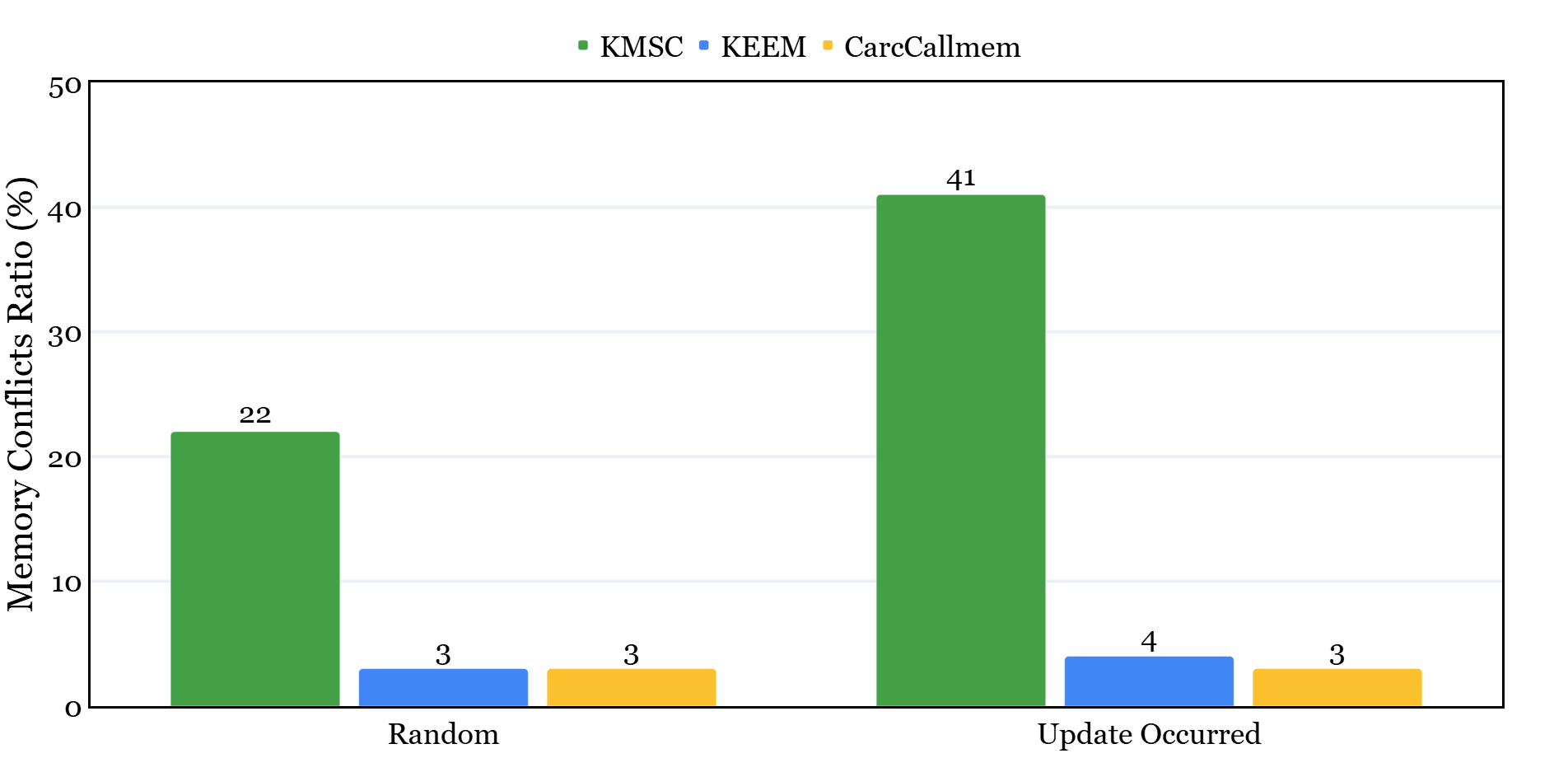}
\caption{Ratio of conflicts between memory sentences across the different memory update methodologies.}
\label{tab_conflicts}
\end{figure*}
% In this section, we discuss the informativeness of updated memory and conflicts between memory sentences.
% To compare the informativeness of summaries and updated memories across the KMSC, KEEM, and \textit{CareCallmem} datasets, we extract keywords from entire session dialogues, $K^{D} = [keyword^{d}_{1}, keyword^{d}_{2}, \dots,  keyword^{d}_{I}]$ and last session's memory, $K^{M} = [keyword^{d}_{1}, keyword^{d}_{2}, \dots, keyword^{d}_{J}]$. Where $I$ and $J$ denote the number of keywords from the whole dialogue and the last session's memory, respectively. After extracting keywords, we analyzed how much $K^{M}$ covers $K^{D}$. 
% GPT revised
In this section, we discuss the informativeness of updated memory and conflicts between memory sentences.
% \begin{table}[t]
% \centering
% \small
% \begin{tabular}{cccc}
% \toprule
% \multirow{2}{*}{\# of keywords} & \multicolumn{3}{c}{Recall}                                                  \\ \cline{2-4}
%                                 & \multicolumn{1}{l}{\small{KMSC}} & \multicolumn{1}{l}{\small{KEEM}} & \multicolumn{1}{l}{\small{CareCallmem}} \\ \midrule
% 5                               & 96\%                     & 97\%                    & 65\%                            \\
% 7                               & 92\%                     & 96\%                     & 63\%                            \\
% 10                              & 87\%                     & 91\%                     & 58\%                           \\ \bottomrule
% \end{tabular}
% \caption{Results of keyword coverage across the different memory update methodologies.}
% \label{tab_keyword_coverage}
% \end{table}
\paragraph{Informativeness of Updated Memory} %\textcolor{red}{요약 논문 레퍼런스 내용 추가!} 
To compare the informativeness of summaries and updated memories across different methodologies—accumulation (i.e., KMSC ), generation (i.e., our KEEM), and operation (i.e., \textit{CareCallmem})—we extract keywords from the entire session dialogues, denoted as $K^{D} = [keyword^{d}_{1}, keyword^{d}_{2}, \dots, keyword^{d}_{I}]$, and from updated memory, denoted as $K^{M} = [keyword^{m}_{1}, keyword^{m}_{2}, \dots, keyword^{m}_{J}]$. Here, $I$ and $J$ represent the number of keywords from the whole dialogue and the last updated memory, respectively. We then analyze the extent to which $K^{M}$ covers $K^{D}$, as follows:
\begin{equation}
    \text{Recall} = \frac{\text{Number of correct keywords in } K^{M}}{\text{Number of keywords in } K^{D}}
\end{equation}
% We used TextRank \cite{mihalcea2004textrank} to extract keywords from the entire session dialogues and the last updated memory with the TF-IDF method. First, we calculated the co-occurrence using all of the words in the whole dialogue. Second, we extracted various numbers of keywords, 5, 7, and 10 from them, then we calculated the recall of keywords. Table \ref{tab_keyword_coverage} shows the results of keyword recall according to the different memory update methods. As shown in Table \ref{tab_keyword_coverage}, the update memories by our generation method, KEEM, outperformed keyword recall in both accumulation and operation-based methods. Indeed, we expected the accumulation method, KMSC, would show the highest keyword recall. Interestingly, our KEEM method shows superior keyword recall compared to other methods. We analyzed these underscores of the KMSC method and concluded that it is attributed to the quality of summaries in the KMSC dataset. Even though the KMSC dataset accumulated their session's summary, it implies that there missing important information in the summary of the KMSC dataset. On the other hand, the operation-based method, \textit{CareCallmem}, showed poor performance of keyword recall. It is due to the \textit{REPLACE} and \textit{DELETE} operations eliminating the memorable information when updating memory. As a result, these higher scores of keyword recall of our KEEM memory demonstrate that updated memory in our KEEM dataset is highly effective and useful.
% GPT revised
We utilize TextRank \cite{mihalcea2004textrank} along with the TF-IDF method to extract keywords from the entire session dialogues and the last updated memory. Initially, we calculate word co-occurrence across the entire dialogue. Subsequently, we extract different sets of keywords, specifically 5, 7, and 10, and calculate their recall. Figure \ref{tab_keyword_coverage} presents the results of keyword recall for different memory update methods. The results demonstrate that the updated memories by our generation method, KEEM, outperform those from both accumulation and operation-based methods in keyword recall. Although we anticipate that the accumulation method, KMSC, would show the highest keyword recall due to its comprehensive memory storage, our KEEM method exhibits superior recall. Upon further analysis, the underperformance of the KMSC method is attributed to the quality of its summaries. Despite the KMSC dataset accumulating session summaries, it appears that important information is missing from these summaries. Meanwhile, the operation-based method, \textit{CareCallmem}, exhibits poor keyword recall performance. This deficiency can be attributed to the \textsc{REPLACE} and \textsc{DELETE} operations, which tend to eliminate essential information during the memory updating process.  Consequently, the higher keyword recall scores of our KEEM dataset confirm that our updated memory is highly effective and useful.

\paragraph{Conflicts between Memory Sentences}
To analyze conflicts between memory sentences according to the update methodologies, we employ the natural language inference (NLI) task. Specifically, we instruct ChatGPT 4.0 to determine whether a hypothesis is an entailment, neutral, or contradiction for all possible sentence pairs in the updated memory, with one memory sentence as the premise and another as the hypothesis. For a thorough analysis of conflicts, we utilize two data sampling methods: first, we randomly sample 50 updated memories; second, we specifically select 50 samples where a memory update occurs. Figure \ref{tab_conflicts} presents the results regarding the sampling method for all datasets. The accumulation method (i.e., KMSC) exhibits the highest ratio of conflicts between memory sentences, likely due to the method's practice of merely accumulating session summaries as conversations progress. In samples with memory updated, the ratio of conflicts significantly increases under the accumulation method, indicating that merely accumulating session summaries is not effective for long-term conversation tasks. For the generation (i.e., KEEM) and operation-based (i.e., \textit{CareCallmem}) methods, both show a significantly lower ratio of conflicts. The operation-based method, due to its \textsc{REPLACE} and \textsc{DELETE} operations, already eliminates sentence pairs that could lead to contradictions. Although this characteristic of the \textit{CareCallmem} method reduces conflicts, it also tends to remove essential information from dialogues. Conversely, our generation-based update method, KEEM, not only effectively retains crucial information in dialogues but also ensures minimal conflicts between memory sentences.

%%% Trade-off Figure 추가하기 %%%
\subsubsection{Evaluation with Long-term Conversation Models}
\begin{table*}[t]
\centering
\begin{tabular}{lcccccc}
\toprule
\multirow{2}{*}{Model}                                             & \multicolumn{3}{c}{Perplexity} & \multicolumn{3}{c}{Voting} \\ \cline{2-7} \\[-0.7em]
                                                                   & KMSC   & KEEM   & CareCallmem  & KMSC  & KEEM & CareCallmem \\ \midrule
RAG                                                                & 9.74   & 8.10   & 10.55        & 26\%  & 67\% & 7\%         \\
FiD                                                                & 9.51   & 7.88   & 10.49        & 22\%  & 68\% & 10\%        \\
FiD-RAG                                                            & 9.66   & 7.90   & 10.52        & 19\%  & 77\% & 4\%         \\ \midrule
Llama2 7B                                                          & 8.70   & 6.99   & 10.50        & 16\%  & 81\% & 3\%         \\
Llama2 13B                                                         & 8.53   & 6.89   & 10.51        & 15\%  & 82\% & 3\%         \\ \midrule
\begin{tabular}[c]{@{}l@{}}Llama10.8B tuned Korean\end{tabular} & 5.47   & 4.56   & 8.13         & 22\%  & 74\% & 4\%         \\ \hline \\[-0.8em]
\begin{tabular}[c]{@{}l@{}}Phi2.8B tuned Korean\end{tabular}    & 5.83   & 4.61   & 8.72         & 18\%  & 79\% & 3\%        \\ \bottomrule
\end{tabular}
\caption{Perplexity results on various models}
\label{tab_perplexity}
\end{table*}
% To assess the application of memories regarding the update methodologies, we utilized various long-term conversation models. In the previous study of multi-session chat, \citet{xu2022beyond} conducted modeling multi-session chat using various models such as RAG (Retrieval-Augmented Generation) \cite{lewis2020retrieval}, FiD (Fusion-in-Decoder) \cite{izacard2021leveraging}, and FiD-RAG \cite{shuster2021retrieval}. They evaluated the ability of multi-session conversation with their proposed, MSC dataset. However, assessing the application of updated memories remains a huge challenge since simply conversing using these models may not cover this aspect as it does not guarantee that information has been updated even in multi-session conversations. To accurately evaluate updated memory, we need a dialogue that converses about updated information. So, we create the fifth session dialogue that ensures having at least one turn talking about updated information. Finally, we assess the application of updated memories from different methodologies by inputting them with a fifth session dialogue into the models, to respond to mentioning updated information utterance.
% GPT revised
To assess the application of updated memories across different methodologies, we utilize various long-term conversation models. Previous research of multi-session chat by \citet{xu2022beyond} employs models such as RAG (Retrieval-Augmented Generation) \cite{lewis2020retrieval}, FiD (Fusion-in-Decoder) \cite{izacard2021leveraging}, and FiD-RAG \cite{shuster2021retrieval} to model multi-session chats, evaluating their performance on the MSC dataset. However, the challenge remains in assessing the application of updated memories, as merely conversing using these models does not ensure that information has been updated, even in multi-session conversations. To address this, we design a fifth session dialogue specifically to include at least one turn discussing updated information, ensuring the dialogue directly engages with the updated memories. We then input this fifth session dialogue into the models and evaluate their ability to respond appropriately to utterances mentioning the updated information.
Following the previous work \cite{xu2022beyond}, we use RAG, FiD, and FiD-RAG. Additionally, we use large language models (LLMs) such as Llama2 \footnote{https://huggingface.co/meta-llama/Llama-2-7b-chat-hf} \footnote{https://huggingface.co/meta-llama/Llama-2-13b-chat-hf}  \cite{touvron2023llama} to assess the application of memories. We also use models tuned with Korean based on Llama\footnote{https://huggingface.co/yanolja/EEVE-Korean-10.8B-v1.0} or Phi-2\footnote{https://huggingface.co/yanolja/EEVE-Korean-2.8B-v1.0}  \cite{li2023textbooks}.
\begin{table*}[t]
\centering
\small
\begin{tabular}{lll}
\toprule
Previous Memory                                                                                                            & Subsequent Session's Summary                                                                               & Updated Memory                                                                                                                                                                                       \\ \midrule
\begin{tabular}[c]{@{}l@{}}My son is a high school student and\\ his grades were good until his first year.\end{tabular}   & \begin{tabular}[c]{@{}l@{}}My son entered the physics department \\ at a university in Seoul.\end{tabular} & \begin{tabular}[c]{@{}l@{}}\textcolor{red}{My son is a high school student}; \\ his grades were good until his first year, \\ and now he has entered the physics\\ department at a university in Seoul.\end{tabular} \\ \midrule
\begin{tabular}[c]{@{}l@{}}I recently visited Yeosu and have chosen \\Damyang as my next travel destination.\end{tabular} & I have been to Jeonju.                                                                                     & \begin{tabular}[c]{@{}l@{}}I recently visited Yeosu and \textcolor{red}{have chosen}\\ \textcolor{red}{Jeonju as my next travel destination.} \\
\end{tabular}                                                                    
\\ \bottomrule
\end{tabular}
\caption{Examples of memory update errors}
\label{examples_errors}
\end{table*}
For the automatic evaluation metric, we utilize perplexity, anticipating that the most effective memory would exhibit the lowest perplexity, indicating lower generation difficulty for the given input. Initially, we input each updated memory from different methodologies (Note: memories are generated using the methodology of \textit{CareCallmem}, as mentioned in Section \ref{sec_automatic_human_eval}) along with a fifth session dialogue into the models. We then evaluate their perplexity when responding to utterances that mentioned the updated information. Subsequently, we assess the responses generated by the models by instructing ChatGPT 4.0 to vote for the most appropriate response given the context of the entire session's dialogue. Table \ref{tab_perplexity} presents the perplexity performance of various models, showing that models tuned with Korean exhibit better perplexity compared to others. Memories updated using the \textit{CareCallmem} method exhibit the highest perplexity, suggesting that the operation-based method is inadequate for retaining essential information, leading to poor response generation. In contrast, the accumulation method (KMSC) displays better perplexity than the operation-based method, despite potential contradictions between memory sentences, as it does not remove memorable information. Our generation-based method, KEEM, outperforms all baselines across various models. These results suggest that our updated memories are more useful and that the KEEM dataset is more effective in managing the recent state or information of the user while retaining memorable information.

\begin{figure*}[t]
\centering
\includegraphics[width=\textwidth]{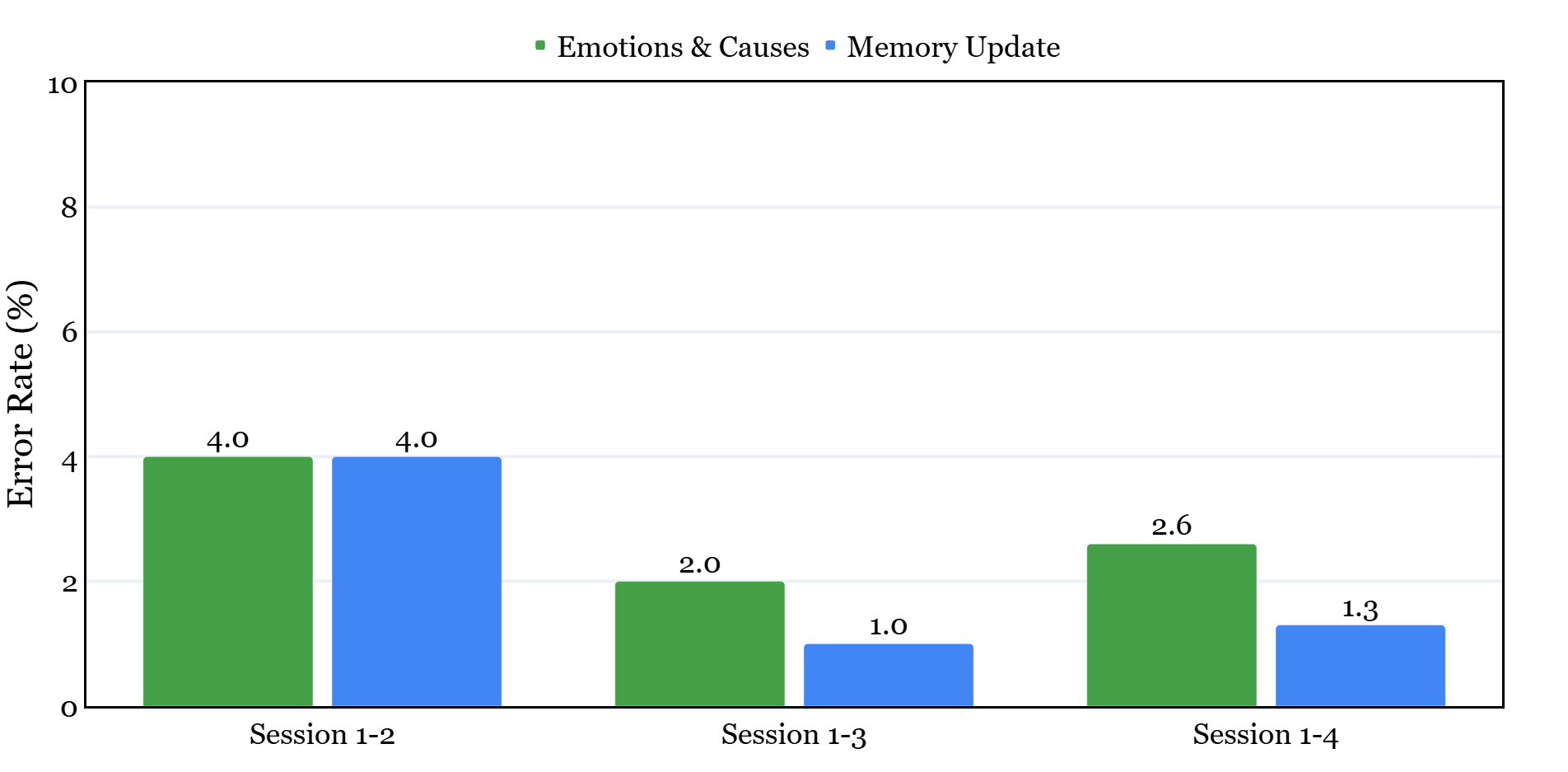}
\caption{Results of generation error analysis.}
\label{tab_generation_error}
\end{figure*}
\section{Analysis}
Because we utilize the ChatGPT API to create the KEEM dataset, we analyze whether hallucinations occurred in the updated memory. In the memory updating process, we instruct ChatGPT to generate updated memory by integrating the previous memory with a new session summary. Consequently, hallucinations—generated content not present in the dialogue—might not occur. We also manually assess these hallucinations and confirm their absence in the KEEM dataset. Therefore, we report the generation errors—such as incorrect integration of unrelated information or contradictions during the integration of related information—that occurred during the memory update process. Table \ref{examples_errors} presents examples of these generation errors. The first example in Table \ref{examples_errors} illustrates a contradiction in the updated memory, while the second example highlights the integration of unrelated information. \\
\indent To assess generation errors, we manually evaluate updated memories in our KEEM dataset across 150 samples, distributed evenly with 50 memories from each of Session 2, Session 3, and Session 4 data. The KEEM dataset creation involves two phases: 1) Emotion \& Causes Reflection (Section \ref{emotion_causes}); and 2) Memory Update (Section \ref{Sec_memory_update}). Generation errors can occur in either phase. To comprehensively analyze these errors, we identify the specific phase in which each error occurs. Figure \ref{tab_generation_error} displays the generation error rates, where Session $n$ indicates data comprising $n$ sessions. The result reveals significantly low generation errors in both phases. Additionally, we sample 70 datasets where information is updated (note that information updates do not always occur) and manually evaluate the occurrence of generation errors during the update process. Only 6 samples exhibited errors, approximately 8.5\%. While this rate of generation error in our KEEM dataset is low, a comparison with other datasets is needed to determine its relative performance. \citet{wang2023selfinstruct} reported error rates of their created dataset in the task description, input, and output generation using GPT-3 \cite{brown2020language} as 8\%, 21\%, and 42\%, respectively, resulting in only 54\% valid data across all fields in their dataset. This underscores the effectiveness of ChatGPT 4.0 in creating new datasets, as evidenced by the notably lower error rate in our KEEM dataset. Moreover, studies on automatic summarization have reported generation error rates around 30\%. Compared to these levels, our KEEM dataset demonstrates a remarkably low error incidence, suggesting a high level of data quality.

\section{Conclusion}
% We introduce the KEEM dataset which is generation-based and integrates both emotion and causality into memory updates for long-term conversation. We demonstrated that our updated memory not only reflects emotions and their causes but also accurately updates user information. Moreover, our memory update method significantly outperformed the previous memory update method which is operation-based that can not retain the user's important information. For future work, we will study the memory update model that can consider the information of time gap, using large language models.
% GPT revised version
% We introduce the KEEM dataset, a generation-based approach that integrates both emotion and causality into memory updates for long-term conversation systems. Our evaluations demonstrate that the updated memory effectively reflects emotions and their causes while accurately updating user information. Furthermore, our method significantly outperforms previous operation-based memory update methods, which often fail to retain crucial user information. For future work, we plan to explore memory update models that account for the temporal gap or terms between sessions, leveraging large language models to enhance the system's memory understanding.
We introduced the KEEM dataset, which integrates emotion and causality into memory updates for long-term conversations. Our evaluations showed that KEEM effectively captures emotions and their causes, outperforming traditional operation-based methods in updating user information. Future work will explore models that consider temporal gaps between sessions, utilizing large language models to improve memory understanding.
\section*{Limitations}
% (1) ChatGPT가 가끔 메모리 내용을 업데이트와 무관하게 삭제하는 경우,
% (2) 메모리 업데이트 데이터를 만들 때 이전 메모리와 현재 메모리만 입력하고 대화를 입력하지 않은 것,
% (3) 생성 방법이기 때문에 hallucination이 포함될 수 있다는 것
% 1) In our analysis, we observe that ChatGPT sometimes deletes some content which is even an unrelated memory update in the memory. To alleviate this problem, we additionally instruct ChatGPT not to delete any content unrelated to memory updates. This method alleviated this problem but did not solve it perfectly. 2) For cost saving, in the memory update process, we input to ChatGPT the previous memory and the current session's summary rather than the current session's dialogue. It could miss important information in dialogue when the current session's summary is not quite good. 3) Because we used ChatGPT which is a generation model, hallucinations could be in the updated memory. To overcome this problem we verify the updated memory again using ChatGPT.
% GPT revised version
1) In our analysis, we noted instances where ChatGPT inadvertently deleted content unrelated to the targeted memory updates, which contradicts the intended behavior of preserving relevant information. To address this issue, we explicitly instructed ChatGPT to avoid deleting any content not directly related to the updates. Although this strategy somewhat mitigated the problem, it did not fully resolve the inadvertent deletion of unrelated content. 2) Furthermore, to reduce costs in the memory update process, we opted to provide ChatGPT with the previous memory and the current session's summary instead of the full dialogue from the current session. This strategy, though cost-effective, risks omitting crucial information if the current session's summary lacks comprehensiveness. 3) Additionally, since ChatGPT is a generative model, there is a risk of generation error being introduced into the updated memory. To address this, we implemented a verification step, re-evaluating the updated memory with ChatGPT to ensure its accuracy and relevance.
\section*{Acknowledgement}
This work was supported by Institute for Information \& communications Technology Promotion (IITP) grant funded by the Korea government (MSIT) (RS-2019-II190004, Development of semisupervised learning language intelligence technology and Korean tutoring service for foreigners). This work was also supported by Institute for Information \& communications Technology Promotion (IITP) grant funded by the Korea government (MSIT) (RS-2024-00398115, Research on the reliability and coherence of outcomes produced by Generative AI).

\bibliography{anthology,custom}

\appendix
\section{Appendix}
\subsection{Details for Topics} 
\label{sec_topic}

\begin{table}[h]
\centering
\begin{tabular}{lcc}
\toprule
Topic                         & Count  & Ratio   \\
\midrule
Individuals \& Relationships & 12,783 & 15.98\% \\
Entertainment                 & 11,063 & 13.83\% \\
Beauty \& Health             & 9,038  & 11.30\%  \\
Society                       & 7,328  & 9.17\%  \\
Work \& Job                  & 6,997  & 8.75\%  \\
Arts \& Culture              & 6,508  & 8.13\%  \\
Education                     & 6,288  & 7.86\%  \\
Food                          & 4,130  & 5.15\%  \\
Climate                       & 3,975  & 4.97\%  \\
Traffic                       & 3,795  & 4.75\%  \\
House                         & 3,489  & 4.37\%  \\
Fashion                       & 235    & 0.29\% \\
\bottomrule
\end{tabular}
\caption{Topic statistics of KMSC dataset}
\label{tab_topic}
\end{table}

Table \ref{tab_topic} shows the various topics used in our KMSC dataset and their statistics.

\subsection{\textit{CareCallmem} Examples \label{Carecallmem_examples}}

\begin{table*}[t]
\centering
\small
\begin{tabular}{l}
\toprule
Input dialogue and summary (translated into English)
\\ \midrule
\textbf{Dialogue:} \\
system: Hello, this is the Administration Welfare Center calling to check in with you.\\
user: Yes.\\
system: Alright, is it okay to talk now?\\
user: Yes.\\
system: How have you been lately?\\
user: I've been tending to an additional houseplant that I got.\\
system: Wow, what kind of plant is it?\\
user: I'm not sure. My husband brought it as a gift.\\
system: I see. Take good care of it.\\
user: Thank you.\\
system: Alright, is there anything bothering you by any chance?\\
user: Nope, nothing at all.\\
system: That's good to hear. I hope you stay healthy in the future too.\\
user: Okay.\\
system: How are you feeling today?\\
user: I'm feeling really good.\\
system: Oh, I see. What were you doing before our call?\\
user: I was cleaning the new houseplant that arrived.\\
system: I see. Having a new plant must make you feel good.\\
user: Yes, it does.\\
system: Keep taking good care of it.\\
user: Yes, thank you.\\
system: Alright, can we end the call here today?\\
user: Sure.\\
system: Take care. Thank you for taking the call.\\
\\
\textbf{Memory:} \\
\begin{tabular}{@{}l@{}}
    · Health condition is good.\\
    · Frequently drinks warm water and tea for health.\\
    · Sleeps well\\
    · Eats rice cakes\\
    · Eats meals well\\
\end{tabular}
\\ \\
\textbf{Summary:} \\
· Has a husband\\
\bottomrule
\end{tabular}
\caption{The example of CareCallmem.}
\label{tab:carecallmem_example}
\end{table*}

Table \ref{tab:carecallmem_example} displays the examples of \textit{CareCallmem} dataset.

\begin{table*}[t]
\centering
\begin{tabular}{l}
\toprule
Instruction (translated into English) \\
\midrule
If there are any inaccuracies or deficiencies in the sentences of {[}\{speaker\}'s summary{]} \\ after understanding the overall flow and content of the {[}conversation{]}, please correct those sentences.\\ In particular, If an emotion is expressed without an explicit cause, \\ please reflect and incorporate the cause of that emotion from the conversation when making corrections.\\ Do not add emotions if the speaker did not explicitly express them.\\ Please note that you should not generate any new content not present in the conversation, \\ and during this process, no new sentences should be added to {[}\{speaker\}'s summary{]}, \\ nor should any existing sentences be deleted.\\ Additionally, {[}\{speaker1\}'s summary{]} should not contain the speaker2's speech.\\ Output the corrected sentences along with the remaining sentences in {[}\{speaker\}'s summary{]} \\ in the same format as {[}Modified \{speaker\}'s summary{]}.\\
\bottomrule
\end{tabular}
\caption{Our prompts that allow the addition of context and causes when only emotions are expressed in summaries of each speaker, enabling understanding of the conversation's content and context.}
\label{table_emotion_instruction}
\end{table*}

\subsection{Emotions \& Causes Reflection Result Examples} \label{appendix:emotions & causes reflection}

\begin{table*}[t]
\centering
\begin{tabular}{l}
\midrule
Input dialogue and summary (translated into English)
\\ \midrule
\textbf{Dialogue:} \\
Speaker1: Hello, I'm a female CEO in my 40s at a mid-sized company.\\
Speaker2: Hi there, I'm a 40-something woman working in a professional field.\\
Speaker1: Nice to meet you. I play VR games for work every day.\\
Speaker2: Oh, I see. I often do yoga after work.\\
Speaker1: Is yoga your hobby? Mine is occasionally playing the oboe.\\ \qquad \qquad It's nice not having neighbors around for this.\\
Speaker2: That's a cool! I sometimes listen to K-pop music, and these days,\\ \qquad \qquad idols are really talented, especially in singing.\\
Speaker1: Absolutely! Nowadays, idols are so versatile.\\ \qquad \qquad Lately, I've been eating fruit to boost my vitamins, but I don't like persimmons. \\ \qquad \qquad Do you like fruit?\\
Speaker2: I'm not really into fruits. \\ \qquad \qquad Lately, I've been feeling a bit restless at times. I've heard that kiwi can help with sleep,\\ \qquad \qquad so I've been making sure to eat plenty of kiwis as I've been having trouble sleeping.\\
Speaker1: I didn't know kiwi could help with sleep! What's been going on with you lately? \\
Speaker2: Well, my brother is a hairstylist, and he keeps talking about expanding his salon, \\ \qquad \qquad especially in times like these...you know, during this economic downturn. \\
Speaker1: It's probably best not to expand recklessly at times like this, you know. \\
Speaker2: I agree. It's important to manage what he have well, but I'm worried. \\
Speaker1: I hope your brother's salon continues to do well! \\
\\
\textbf{Summary:} \\
{[}speaker2's summary{]} \\
\begin{tabular}{@{}l@{}}
    · I am a woman in my 40s working in a professional field. \\
    · I often do yoga after work. \\
    · I listen to K-pop music occasionally. \\
    · I don't particularly like fruit, but I make sure to eat kiwi. \\
    · I've been feeling anxious occasionally lately. \\
    · My brother is a hairdresser. \\
    · I'm worried about my brother. \\
\end{tabular}
\\ \\
{[}modified speaker2's summary{]}\\ 
\begin{tabular}{@{}l@{}}
    · I am a woman in my 40s working in a professional field. \\
    · I often do yoga after work. \\
    · I listen to K-pop music occasionally. \\
    · I don't particularly like fruit, but I make sure to eat kiwi because it helps with sleep. \\
    · I sometimes feel anxious and have trouble sleeping because my brother is considering \\  \quad expanding his hair salon during the recession, which worries me. \\
    · My brother is a hairdresser. \\
    · I'm worried about my brother. \\
 \end{tabular}
 \\
\bottomrule
\end{tabular}
\caption{The example \#1 of input dialogue and its summary. The modified summary reflects the emotion and its cause.}
\label{summary_example_emotion_prompt_part1}
\end{table*}

Table \ref{summary_example_emotion_prompt_part1}, \ref{summary_example_emotion_prompt_part2} show the examples of emotion and its cause reflection.

% \subsection{Data Statistics for \textit{KMSC} and KEEM \label{statistics_datasets}}

\begin{table*}[t]
\centering
\begin{tabular}{l}
\midrule
Input dialogue and summary (translated into English)
\\ \midrule
\textbf{Dialogue:} \\
Speaker1: It's been 4 weeks! Have you been doing well? I wonder if you went on a trip \\
Speaker2: Yeah, I had a great trip. I had a blast after a long time!\\ \qquad \qquad \ How have you been for the past month?  \\
Speaker1: I've been good. How was your trip to Gangneung? Was it as good as you expected? \\
Speaker2: Yeah! It was so much fun! I even had lobster, and it was both reasonably priced \\ \qquad \qquad \ and delicious. How's your pet doing? \\
Speaker1: I'm doing well! I'm planning to go back to the vet today after about a month. \\ \qquad \qquad Feeling a bit worried. \\
Speaker2: Oh, if you're worried, now I'm worried too... Was it usual for you to go once a month? \\
Speaker1: I went because I've been feeling sick recently. I got treated for it, \\ \qquad \qquad \ but I still want to make the visit. Is it alright? \\
Speaker2: I'm worried to hear that you've been feeling sick. Since you've got treated,\\ \qquad \qquad \ everything should be fine. Don't worry too much. \\
Speaker1: It's not good to worry too much, right? I guess I'll just try to clear my mind \\ \qquad \qquad \ while having a meal.. Have you had your meal? \\
Speaker2: I haven't eaten yet! I'm about to eat now. What kind of food are you thinking of having? \\
Speaker1: I'm planning to have some soft beef soup! Do you happen to like beef soup? \\
Speaker2: Yeah, I like various types of soups. Today, I suddenly felt like having pork cutlet,  \\ \qquad \qquad \ so I'm planning to have that! \\
Speaker1: Pork cutlet sounds good! Well then, enjoy your meal, and let's talk again next time~ \\
Speaker2: Sure thing! Enjoy your meal, and have a good visit to the hospital! See you next time! \\
\\
\textbf{Summary:} \\
{[}speaker1's summary{]} \\
\begin{tabular}{@{}l@{}}
· I've been doing well. \\
· I'm going to go back to the animal hospital again. \\
· I'm going to try to clear my mind while having a meal. \\
· I'm going to have some soft beef soup. \\
\end{tabular}
\\ \\
{[}modified speaker1's summary{]}\\ 
\begin{tabular}{@{}l@{}}
· I'm planning to go back to the animal hospital because I'm worried that my pet might be sick. \\
· I think it's not good to worry too much, so I'm going to try to clear my mind while having a meal. \\
· I'm going to have some soft beef soup. \\
\end{tabular}
\\
\bottomrule
\end{tabular}
\caption{Example of input for Dialogue and Summary. The modified summary reflects the emotion and its cause.}
\label{summary_example_emotion_prompt_part2}
\end{table*}

\subsection{Memory Update Result Examples \label{Appendix_memory_update_examples}}
Table \ref{example_memory_update_result} shows our memory update instruction and its result. Table \ref{example_memory_update_result_part1}, \ref{example_memory_update_result_part2} show the used few-shot examples for memory updates.

\begin{table*}[t]
\centering
\begin{tabular}{l}
\midrule
Input dialogue and summary (translated into English)
\\ \midrule
\#\#\# Instruction: You are a language model that operates in a multi-session chatbot. In cases where\\ \qquad the information in {[}\{speaker\}'s summary{]} necessitates the modification or deletion of a sentence in \\ \qquad  {[}Memory{]}, you update the sentence to reflect the latest information while maintaining \\ \qquad consistency. The sentences in {[}Memory{]} are accumulated from past conversations, whereas \\ \qquad {[}\{speaker\}'s summary{]} reflects the current conversation.\\ \qquad Therefore, I will acknowledge that {[}\{speaker\}'s summary{]} contains more recent information\\ \qquad  compared to {[}Memory{]} and update accordingly. You won't merge or modify sentences \\ \qquad in {[}Memory{]}  just because they're on the same topic. You'll only update sentences in {[}Memory{]}\\ \qquad when they need to be changed based on the content of {[}\{speaker\}'s summary{]} or \\ \qquad when the content continues. \\ \qquad When updating, ensure that no new content is generated, and existing content is not lost. \\ \qquad Referencing examples 1 and 2, merge the updated {[}Memory{]} and {[}\{speaker\}'s summary{]}\\ \qquad to produce the output in example 3. \\
 \\
\#\#\# Input: \\
\qquad Example1: \\
\qquad \{Example 1\}\\
\\
\qquad Example2: \\
\qquad \{Example 2\}\\
 \\
\#\#\# Output: \\
\qquad Example3: \\
\qquad {[}Memory{]} \\
\qquad  I am a woman in my forties. \\
\qquad  I live in Seoul. \\
\qquad I am a traditional Korean medicine doctor. \\
\qquad I am currently writing a paper on Oriental acupuncture. \\
\qquad It's been just under a month since I started writing the paper. \\
\qquad I'm ambitious and want to make my name known. \\
\qquad I feel jealous when my friends are more successful than me. \\
\qquad I have been commuting by public transportation for 10 years, and it's exhausting. \\
\\
\qquad {[}\{speaker\}'s summary{]} \\
\qquad I've been busy writing my paper lately, but I've just finished writing it. \\
\\
\qquad {[}Updated memory{]} \\
\qquad I am a woman in my forties. \\
\qquad I live in Seoul. \\
\qquad I am a traditional Korean medicine doctor. \\
\qquad I've been busy writing a paper on Oriental acupuncture lately, and I've just finished writing it. \\
\qquad I am ambitious and want to make my name more known. \\
\qquad I feel jealous when my friends are more successful than me. \\
\qquad I have been commuting by public transportation for 10 years, and commuting is tough. \\ \\
\bottomrule
\end{tabular}
\caption{Example of Memory update instruction}
\label{example_memory_update_result}
\end{table*}

\begin{table*}[t]
\centering
\begin{tabular}{l}
\midrule
Few-shot examples used for memory updates(translated into English)
\\ \midrule
{[}Memory{]} \\ 
I am waiting for my car to be repaired.\\
I work at a trading company.\\
I am a high school student living in Yangcheon-gu.\\
I definitely want to visit Universal Studios if I go to Japan.\\
I haven't booked a flight to Japan yet.\\
I haven’t reconciled with my boyfriend yet.\\
I’ve developed an interest in interior design and have been looking at furniture lately.\\
I’m working overtime.\\
I’m having doubts about marrying my boyfriend.\\ \\
{[}Summary{]} \\ 

My car was ultimately declared a total loss.\\
I quit my job and became a housewife after having a child.\\
I find it hard to concentrate when taking video classes.\\
My overtime work is finished.\\
I moved to Guro-gu because of my father’s company.\\
I think the public transportation transfer system is good.\\
I eventually broke up with my boyfriend.\\
I’ve booked a flight and am now looking for accommodation.\\ \\
{[}Updated Memory{]}\\
My car, which I had left for repairs, was ultimately declared a total loss.\\
I worked at a trading company but became a housewife after having a child.\\
I’ve developed an interest in interior design and have been looking at furniture lately.\\
I definitely want to visit Universal Studios if I go to Japan.\\
I used to live in Yangcheon-gu but moved to Gwangjin-gu because of my father's company.\\
I find it hard to concentrate when taking video classes.\\
I eventually broke up with my boyfriend, with whom I had been contemplating marriage.\\
I think the public transportation transfer system is good.\\
I’ve booked a flight to Japan and am currently looking for accommodation.\\
\\
\bottomrule
\end{tabular}
\caption{Few-shot examples used for memory updates (Part 1)}
\label{example_memory_update_result_part1}
\end{table*}

\begin{table*}[t]
\centering
\begin{tabular}{l}
\midrule
Few-shot examples used for memory updates(translated into English)
\\ \midrule
{[}Memory{]} \\ 
I am looking for a part-time job near my home.\\
I buy a diary every year but have never used one properly.\\
I am a woman in my 40s living in Seocho-gu.\\
I’ve always been abroad on my birthday.\\
I am a university student attending Sejong University.\\
I tested positive for COVID-19 and am currently in quarantine.\\
I am a speech therapist.\\
I am planning to do volunteer work next weekend.\\
I plan to volunteer at an animal shelter.\\
I have been volunteering at the same animal shelter for two years.\\ \\
{[}Summary{]} \\ 

I majored in physical education but work as a speech therapist.\\
Yesterday was my graduation ceremony.\\
I completed my volunteer work.\\
I recently moved to Yangcheon-gu.\\
I decided to work morning shifts only at a restaurant near my home.\\
I will start my part-time job next week.\\
I have recovered from COVID-19.\\ \\
{[}Updated Memory{]}\\
I graduated from University.\\
I buy a diary every year but have never used one properly.\\
I am a woman in my 40s who moved from Seocho-gu to Yangcheon-gu.\\
I’ve always been abroad on my birthday.\\
Although I majored in physical education, I work as a speech therapist.\\
I did volunteer work at an animal shelter over the weekend.\\
I have been volunteering at the same animal shelter for two years.\\
Starting next week, I will work part-time only in the mornings at a restaurant near my home.\\
Although I had COVID-19, I have fully recovered.\\
\\
\bottomrule
\end{tabular}
\caption{Few-shot examples used for memory updates (Part 2)}
\label{example_memory_update_result_part2}
\end{table*}

\subsection{Examples of \textit{CareCallmem} Instruction} \label{appendix_carecall_instruction}
\begin{table*}[t]
\centering
\resizebox{\textwidth}{!}{
\begin{tabular}{l}
\midrule
Input dialogue and summary (translated into English)
\\ \midrule
\#\#\# Instruction: You are a model that updates information by combining existing information sentences\\ \qquad from [Memory] with new information sentences from [{speaker}'s summary]. \\ \qquad  In this process, you must ensure that the updated information is consistent, non-redundant, and complete. \\ \qquad The update process follows these steps \\ \qquad 1. For each sentence in [Memory], classify its relationship with each sentence in\\ \qquad  [{speaker}'s summary] using one of four labels: \\ \qquad "PASS" label: The sentence from [Memory] is implied by a sentence in [{speaker}'s summary]. \\ \qquad "REPLACE" label: The sentence from [{speaker}'s summary] \\ \qquad contradicts or implies the sentence in [Memory]. \\ \qquad "APPEND" label: The sentences from [Memory] and [{speaker}'s summary] are neutral towards each other. \\ \qquad "DELETE" label: Both the sentence from [Memory] and [{speaker}'s summary] \\ \qquad are no longer true or no longer need to be remembered. \\ \qquad 2. If a sentence is classified as "REPLACE,"  delete the sentence from [Memory]. \\ \qquad If classified as "DELETE," delete both the sentence from [Memory] and [{speaker}'s summary]. \\ \qquad 3. For each sentence in [{speaker}'s summary], \\ \qquad classify its relationship with each sentence in [Memory] using one of four labels. \\ \qquad 4. If a sentence is classified as "PASS," delete the sentence from [{speaker}'s summary]. \\ \qquad 5. Merge the final versions of [Memory] and [{speaker}'s summary]. \\ \qquad Refer to examples 1 and 2 for guidance, and output the updated result as shown in example 3.
 \\
\#\#\# Input: \\
\qquad Example1: 
\qquad \{Example 1\}\\
\\
\qquad Example2: 
\qquad \{Example 2\}\\
 \\
\#\#\# Output: \\
\qquad Example3:
\quad {[}Memory{]} \\
\qquad  I am looking for a nearby place to work part-time. \,  I buy a planner every year but never use it properly. \\
\qquad I am a woman in my 40s living in Seocho-gu. \, I have always been abroad on my birthday. \\
\qquad I am a university student attending Sejong University. \, I tested positive for COVID-19 and am currently in isolation. \\
\qquad I am a speech therapist. \, I am planning to volunteer next weekend. \\
\qquad I volunteer at an animal shelter. \, I have been volunteering at the same animal shelter for two years.\\ 
\\
\qquad {[}\{speaker\}'s summary{]} \\ \qquad I majored in physical education but work as a speech therapist. \\ \qquad I had my graduation ceremony yesterday. \\ \qquad I recently moved to Yangcheon-gu.\\ I decided to work part-time in the mornings at a nearby restaurant. \\ \qquad I start my part-time job next week.\\ I have recovered from COVID-19.
\\ \\
\qquad {[}Updated{]} \\
\qquad I graduated from Sejong University.
\, I buy a planner every year but never use it properly. \\
\qquad I used to live in Seocho-gu but recently moved to Yangcheon-gu. \\
\qquad I have always been abroad on my birthday. \, I majored in physical education but now work as a speech therapist. \\
\qquad I volunteered at an animal shelter last weekend.
\, I have been volunteering at the same animal shelter for two years. \\
\qquad I will start working part-time in the mornings at a nearby restaurant next week. \, I had COVID-19 but have recovered. \\
\bottomrule
\end{tabular}
}
\caption{Example of memory update instruction using CareCallmem's methodology}
\label{tab:Examples_carecall_instruction}
\end{table*}

Table \ref{tab:Examples_carecall_instruction} shows the instruction of \textit{CareCallmem} method.


\bibliographystyle{acl_natbib}

\end{document}